\PassOptionsToPackage{table}{xcolor}
\documentclass[onecolumn]{fairmeta}

\usepackage{hyperref}

\usepackage{amsmath}
\usepackage{amssymb}
\usepackage{mathtools}
\usepackage{bm}
\usepackage{microtype}
\usepackage{graphicx}
\usepackage{subcaption}
\usepackage{booktabs}
\usepackage{epigraph}
\usepackage{xspace}
\usepackage{xstring}
\usepackage{array}
\usepackage{algorithm}
\usepackage{algorithmic}
\usepackage{amsmath}
\DeclareMathOperator*{\argmax}{arg\,max}
\usepackage{wrapfig}
\usepackage{array}
\usepackage{nicefrac}
\usepackage{amsfonts}
\usepackage{url}

\usepackage{wrapfig} 
\usepackage{lipsum}  

\usepackage{tcolorbox}
\tcbuselibrary{breakable, skins}

\newtcolorbox{rubricbox}{
  enhanced, breakable,
  colback=blue!4,           
  colframe=blue!55!black,   
  coltitle=white,
  fonttitle=\bfseries,
  title={AR-Eval Rubric (judge instruction)},
  boxrule=0.5pt, arc=2pt,
  left=6pt, right=6pt, top=6pt, bottom=6pt,
}
\crefname{table}{Table}{Tables}
\Crefname{table}{Table}{Tables}

\crefname{section}{Section}{Sections}
\Crefname{section}{Section}{Sections}
\crefname{figure}{Figure}{Figures}
\Crefname{figure}{Figure}{Figures}

\usepackage{nicematrix}
\newcommand{\pmSep}{%
  \ifnum\theiRow>2\relax 
    $\,\pm\,$%
  \fi
}
\usepackage{multirow}

\usepackage{tcolorbox}
\tcbuselibrary{skins}
\tcbuselibrary{breakable}
\usepackage{float}
\usepackage{siunitx}
\newcommand{\circa}{{\raise.17ex\hbox{$\scriptstyle\sim$}}}
\usepackage{graphicx} 
\usepackage{booktabs}
\usepackage{pifont}

\usepackage[textsize=tiny]{todonotes}

\usepackage{tikz}
\usetikzlibrary{patterns}

\newtcolorbox{questionbox}[1]{
  colback=metabg,
  colframe=metafg,
  fonttitle=\bfseries,
  title=#1
}

\definecolor{icmlblue}{RGB}{0,76,153} 
\definecolor{icmlgray}{gray}{0.15}
\definecolor{icmllight}{gray}{0.96}

\tcbset{
  icmlbox/.style={
    enhanced,
    breakable,
    boxrule=0.4pt,
    colframe=icmlblue!65!black,
    colback=icmllight,
    coltitle=black,
    fonttitle=\bfseries,
    left=1.0mm,right=1.0mm,top=0.8mm,bottom=0.8mm,
    boxsep=1.2mm,
    arc=1.2mm,
    outer arc=1.2mm,
    titlerule=0pt,
    toptitle=0.4mm,
    bottomtitle=0.2mm,
    attach title to upper,
    title={#1},
  },
  icmlbox/.default={}
}
\newtcolorbox{callout}[1][]{%
  enhanced, breakable, frame hidden,
  left=2mm,right=2mm,top=2mm,bottom=2mm,
  boxrule=1pt,
  colback=metabg,
  arc=10pt,
  before skip=15pt,
  grow to left by=3pt,
  grow to right by=3pt,
  #1
}

\title{ \raisebox{-0.2\height}{\includegraphics[height=0.8\baselineskip]{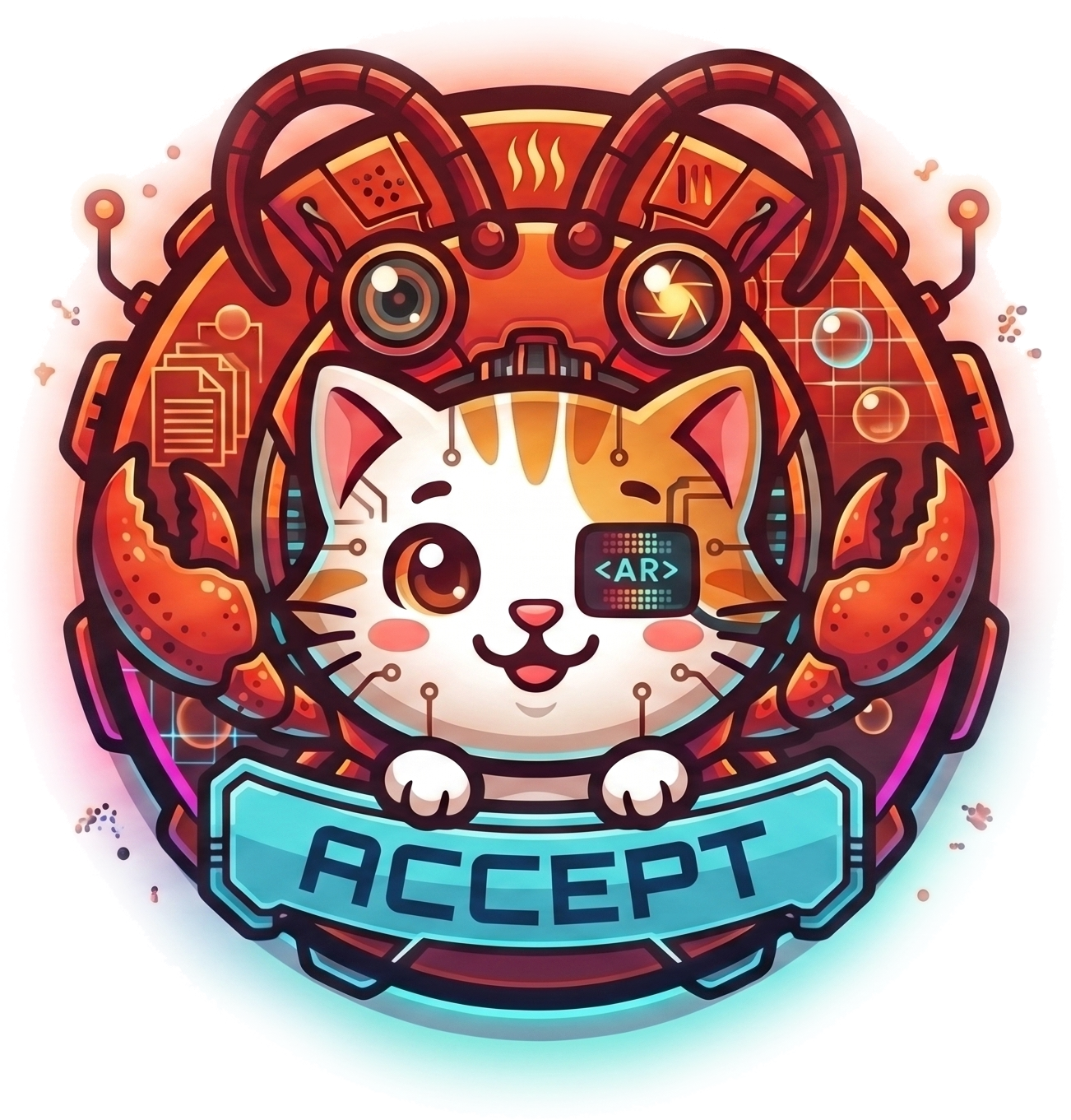}}%
One Reflection Is Not Enough: Self-Correcting Autonomous Research via Multi-Hypothesis Failure Attribution
}

\author[1]{Jie Ma}
\author[2]{Binfei Chu}
\author[1]{Jie Gao}
\author[1]{Jinlu Zhang}
\author[3]{Yiwei Ma}
\author[1]{Yi Tan}
\author[1]{Jiayi Ji}
\author[1]{Xiaoshuai Sun}
\author[1]{Rongrong Ji}

\affiliation[1]{Xiamen University}
\affiliation[2]{Hello Inc.}
\affiliation[3]{Xiaohongshu Inc.}

\abstract{
Autonomous research agents can now draft hypotheses, write code, run experiments, and produce papers, but they remain brittle when experiments fail. Under the prevailing paradigm, failure recovery is usually delegated to a single free-form reflection: a rich trajectory of metrics, logs, and design choices is compressed into one verbal critique, which often leads either to localized trial-and-error or to hard pivots that discard useful context.
We propose \textbf{\textsc{SAGE}}, a \textit{Self-correcting, Autonomous, Grounded Experimenter}, to tackle this failure-recovery bottleneck. Its core mechanism, \textit{Multi-Hypothesis Failure Attribution (MHFA)}, treats recovery as a structured causal diagnosis. By analyzing dynamic trajectory features, MHFA systematically generates multiple evidence-grounded explanations for a failure, independently evaluates their severity, and deterministically routes the verified root cause to the correct intervention level (hypothesis, experimental design, or implementation). To guarantee scientific honesty, \textsc{SAGE} further employs a grounded reporting mechanism that explicitly constrains drafted results to actual measured values, redacting hallucinated numbers.
On a 12-topic, 5-domain benchmark, \textsc{SAGE} increases metrics-bearing outputs from 42\% to 92\% over a reflection baseline, improves artifact quality from 5.00 to 6.75/10, and blindly outscores AI-Scientist-v2 (52.0 vs.\ 48.2), with gains concentrated in code development and execution. While fully autonomous scientific writing and generating conference-ready papers remain notoriously difficult open problems for the entire field, \textsc{SAGE} successfully produces significantly more reliable and higher-quality scientific artifacts. Ultimately, by coupling structured recovery with explicit grounding constraints, \textsc{SAGE} significantly outperforms monolithic reflection paradigms, establishing a highly trustworthy foundation for future autonomous research.
}

\date{\today}
\correspondence{\email{jiema100@stu.xmu.edu.cn},  \email{xssun@xmu.edu.cn}
}
\metadata[Code]{\url{https://github.com/JieMaMagic/SAGE.git}}

\begin{document}

\maketitle

\section{Introduction}
\label{sec:intro}

\begin{figure}[t]
    \centering
    \includegraphics[width=\linewidth]{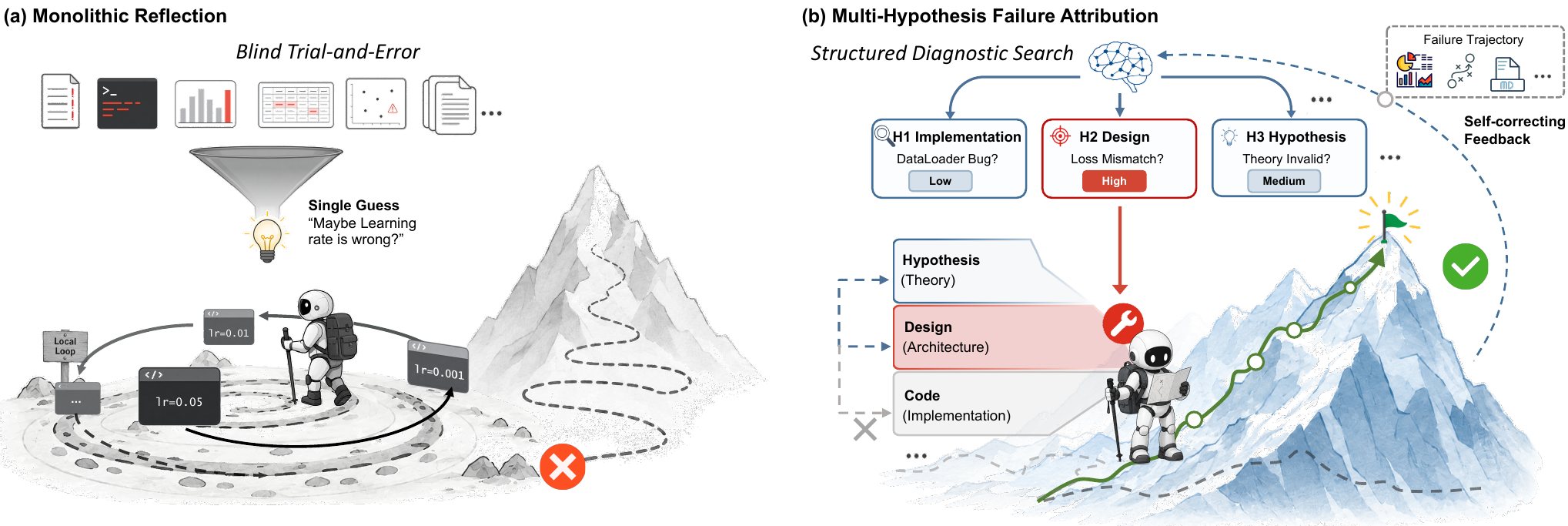} 
    \caption{\textbf{The Failure Recovery Bottleneck.} 
    \textbf{(a) Monolithic Reflection:} Existing agents compress failure logs into a single guess, trapping them in blind, localized trial-and-error. 
    \textbf{(b) \textsc{SAGE} (Ours):} \textsc{SAGE} leverages failure trajectories to generate and rank multiple causal hypotheses. A deterministic router maps the highest-scoring cause to the correct abstraction level (e.g., Design instead of Code). Through structured diagnosis and iterative self-correction, \textsc{SAGE} successfully escapes local optima to rescue deeply flawed experiments.}
    \label{fig:motivation}
\end{figure}

Autonomous research agents are moving beyond tool-use demonstrations toward end-to-end scientific workflows: they can formulate hypotheses, implement experiments, analyze results, and draft papers with limited human intervention~\citep{aiscientist, aide, mlebench, internagent15, airesearcher, alphaevolve, rdagent, deepscientist}. This progress makes an old scientific problem newly central for agents: experiments often fail. A model may plateau, a metric may be insensitive, a protocol may test the wrong claim, or the code may silently emit no usable measurement. In such cases, autonomy depends not only on executing a planned pipeline, but on \emph{navigating failure}: diagnosing the level at which the run broke down and redirecting the research trajectory accordingly~\citep{whyllmsarentscientists, deepresearchsurvey, autogptodm, agentbench, sibylautoresearch, autoresearchclaw, internagent, biomni}.

Existing agents usually recover through monolithic reflection inherited from Reflexion~\citep{reflexion, react} and Self-Refine~\citep{selfrefine, sweagent, openhands, openhandssdk}. This is effective for many self-contained coding tasks, but it is a poor fit for open-ended research because the same symptom can have causes at different abstraction levels. Under a single-attribution, single-action paradigm, agents tend to fall into two failure modes (as shown in \cref{fig:motivation}(a)). They perform localized trial-and-error, such as repeatedly tuning hyperparameters of a fundamentally flawed design~\citep{camyla, mlestar, deepscientist, autoresearchclaw}, or they make a hard pivot that resets the pipeline and discards accumulated evidence~\citep{whyllmsarentscientists, webarena, autogptodm, agentbench}.

The underlying difficulty is \emph{structural credit assignment}: a flat learning curve, degenerate metric, or runtime error can arise from a weak hypothesis, a misaligned evaluation design, or an implementation defect~\citep{sibylautoresearch, storm, paperqa2, openscholar}. An experienced researcher responds by entertaining competing explanations and seeking discriminating evidence, a process formalized as strong inference~\citep{platt1964stronginference}. Monolithic reflection instead compresses this multi-level causal ambiguity into a single verbal summary. The agent may then settle on the most salient, easily editable symptom while missing a deeper methodological problem~\citep{whyllmsarentscientists, tongyideepresearch, webresearcher, deepresearchsurvey}.

We propose \textbf{Multi-Hypothesis Failure Attribution (MHFA)}, a recovery framework that operationalizes the method of multiple working hypotheses~\citep{chamberlin1890, platt1964stronginference} for autonomous research agents. As illustrated in \cref{fig:motivation}(b), MHFA structures recovery as a three-stage process of divergent generation, convergent scoring, and deterministic routing. A divergent generator first synthesizes distinct, evidence-grounded explanations of the failure trajectory. An independent critic then scores the severity and evidential support of each explanation. Finally, rather than letting the model hallucinate a fix, a deterministic routing rule maps the verified structural cause to a hierarchy-aware intervention, such as method refinement, protocol redesign, or hypothesis pivoting. Low-level zero-metric execution failures are handled by a separate explicit repair path before structural attribution. MHFA is orthogonal to multi-path \emph{solution} search, such as Tree-of-Thoughts~\citep{tot} or multi-agent debate~\citep{debate, aris}. Those methods diversify candidate actions or opinions, whereas MHFA diversifies causal \emph{explanations} of an observed failure and routes recovery at the appropriate abstraction level.

We realize MHFA in \textbf{\textsc{SAGE}}, an autonomous research system built on a multi-stage pipeline. To support diagnosis, a \textit{TrajPivot} component supplies trajectory features for attribution. To support faithful reporting, a two-stage numeric grounding mechanism, consisting of a proactive manifest and a strict sanitizer, checks table values against measured result registries and redacts unverified cells. This mechanism is deliberately narrower than full paper verification. It constrains numeric tables, but does not by itself guarantee that prose-level method descriptions, dataset names, libraries, or statistical tests match the executed code.

On a twelve-topic benchmark spanning five domains (machine learning, statistics, quantum computing, biology, and high-energy physics), \textsc{SAGE} improves metrics-bearing recovery from 42\% (5/12 topics) under a reflection baseline to 92\% (11/12 topics). Under a blind, uniform evaluation of complete artifacts, \textsc{SAGE} scores 52.0 out of 100, compared with 48.2 for AI-Scientist-v2 and 24.8 for the host-pipeline reflection baseline, with gains heavily concentrated in code development and execution. Crucially, while generating fully conference-ready papers remains a notoriously difficult open challenge for the entire field, our analysis reveals a fundamental \emph{bottleneck shift}. Once structured recovery renders autonomous experiments reliably executable and empirically measurable, the fundamental limits of autonomy shift to downstream processes. Faithful implementation, prose-table consistency, and method-provenance grounding manifest as the critical open frontiers for future research~\citep{sweagent, openhands, openhandssdk, paperbanana}.

In summary, our contributions are:
\begin{itemize}
\item We reframe agent reflection from a single-chain verbal critique into a \textbf{multi-hypothesis attribution framework}. By structuring causal attribution over failed research trajectories, MHFA preserves competing explanations before committing to a repair.
\item We introduce \textbf{hierarchy-aware recovery routing} to decouple diagnosis from action. A deterministic rule maps verified structural causes to interventions at the appropriate abstraction level, while explicit execution-failure handling repairs zero-metric runs before deeper pivots are attempted.
\item We conduct an \textbf{empirical failure-recovery study} showing that \textsc{SAGE} improves metrics-bearing recovery (11/12 topics vs.\ 5/12) and code-oriented evaluation scores over a reflection baseline, and achieves stronger overall artifact scores than AI-Scientist-v2 in our blind uniform evaluation. Our analysis further identifies method-provenance grounding as a key remaining bottleneck for reliable autonomous research.
\end{itemize}

\section{Related Work}
\label{sec:related}

\paragraph{Autonomous Scientific Discovery.}
Autonomous science has long aimed to close the loop spanning hypothesis generation, experiment execution, evidence interpretation, and subsequent revision~\citep{robotadam,robotfuncgenomics,roboteve,selfdrivingthinfilm}. Earlier robot scientists and self-driving laboratories demonstrated this loop in constrained physical domains, while recent LLM-based systems have expanded it to broader scientific workflows. In physical laboratory settings, agents have been coupled with chemistry tools and robotic execution~\citep{chemcrow,coscientist,organa}, as well as autonomous synthesis platforms~\citep{alab,qiushi,selfdrivingthinfilm}. In digital research, systems such as the AI Scientist, AI Scientist-v2, Agent Laboratory, and the AI Co-Scientist automate increasingly large portions of ideation, implementation, experimentation, review, and hypothesis refinement~\citep{aiscientist,aiscientistv2,agentlaboratory,aicoscientist}. Related frameworks further explore autonomous innovation, evolutionary discovery, and self-reinforcing research pipelines~\citep{airesearcher,alphaevolve,autoresearchclaw,sibylautoresearch}.

These systems expand the \emph{breadth} of autonomous research, but recent evaluations warn that breadth alone does not guarantee reliability. Benchmarks for scientific discovery and ML research reveal persistent failures in experimental execution and result validity~\citep{discoverybench,scienceagentbench,expbench,mlrbench}. Complementary evaluation suites expose weaknesses in scientific coding, paper replication, and real-world software repair~\citep{labbench,scicode,paperbench,swebench}. Audit-oriented studies further show that superficially plausible generated papers can conceal weak evidence, mismatches between plans and execution, fabricated data, paper-artifact mismatches, or premature claims of success~\citep{researcharena,sciintegritybench,whyllmsarentscientists}. Our reporting guard addresses only the numeric-table part of this broader grounding problem. MHFA targets the diagnostic step before control: deciding whether a failed run reflects an invalid hypothesis, a flawed protocol, a weak metric, or an implementation failure.

\paragraph{Reflection and Continual Self-Improvement.}
A central promise of autonomous agents is not merely execution, but continual improvement from experience. Reflection-based methods convert feedback into reusable memories or critiques, enabling agents to revise their behavior without weight updates~\citep{reflexion,selfrefine,critic,selfdebugging}. Longer-horizon systems extend this idea by accumulating skills, execution traces, or procedural knowledge across tasks and trials~\citep{voyager,evoscientist,camyla,datainterpreter}. In parallel, optimization-oriented agents reuse evaluated trajectories to improve prompts, programs, reward functions, or data-science workflows~\citep{opro,eureka,aide}. Together, these works suggest a useful paradigm of \emph{continual learning through reflection}, where failures become inputs for future improvement.

However, existing reflection mechanisms typically extract lessons from failure only after compressing it into a single critique, memory, or repair instruction. This compression is limiting for open-ended autonomous science, where meaningful improvement depends on identifying exactly \emph{what kind} of failure occurred. The same plateaued experiment may indicate a weak scientific hypothesis, an insensitive protocol, a misaligned metric, or a subtle code bug. MHFA makes continual self-improvement more structurally grounded: rather than reducing a failed trajectory to one critique, it expands the trajectory into multiple evidence-backed failure hypotheses, ranks them by severity and support, and routes recovery to the appropriate scientific abstraction level.

\paragraph{From Search to Attribution.}
A related line of work improves reasoning by maintaining multiple candidate paths rather than committing to a single one. Paradigms like Self-Consistency~\citep{selfconsistency}, Tree of Thoughts~\citep{tot}, Graph of Thoughts~\citep{got}, and Multi-Agent Debate diversify reasoning traces~\citep{debate}, thought states, and opinions before selecting or aggregating a final answer. Agentic search methods~\citep{lats,aide,aris,alphaevolve} further combine planning, feedback, and value estimation over actions, code variants, or research trajectories. Hypothesis-oriented systems~\citep{hypothesissearch,aicoscientist,airesearcher} also demonstrate that explicitly generating and testing alternatives can improve inductive reasoning and scientific exploration.

MHFA adopts the principle of maintaining alternatives, but changes what is diversified. Prior work diversifies answers, actions, programs, hypotheses, or solution trajectories; MHFA diversifies \emph{causal explanations of an observed failure}. Inspired by the method of multiple working hypotheses and strong inference~\citep{chamberlin1890,platt1964stronginference}, MHFA reframes recovery as structural credit assignment across the research stack. Rather than directly choosing the next candidate action, it first diagnoses the likely structural bottleneck, then maps that diagnosis to a hierarchy-aware recovery decision, such as hypothesis pivoting, protocol redesign, or method refinement. Low-level zero-metric failures are handled by a separate execution-repair path before structural attribution.

\section{Method}
\label{sec:method}

\textsc{SAGE} (\emph{Self-correcting, Autonomous, Grounded Experimenter}) recovers from failed experiments through a structured, multi-stage pipeline rather than relying on a single free-form reflection. When a run fails, prior autonomous scientists often feed the execution logs back to an LLM, collapsing competing causes into a single, weakly verified guess. \textsc{SAGE} separates this recovery problem into three parts: diagnosing the likely failure level, routing the repair to an appropriate intervention, and constraining numeric reporting to values that can be traced to measured artifacts.

\begin{figure}[t]
    \centering
    \includegraphics[width=\linewidth]{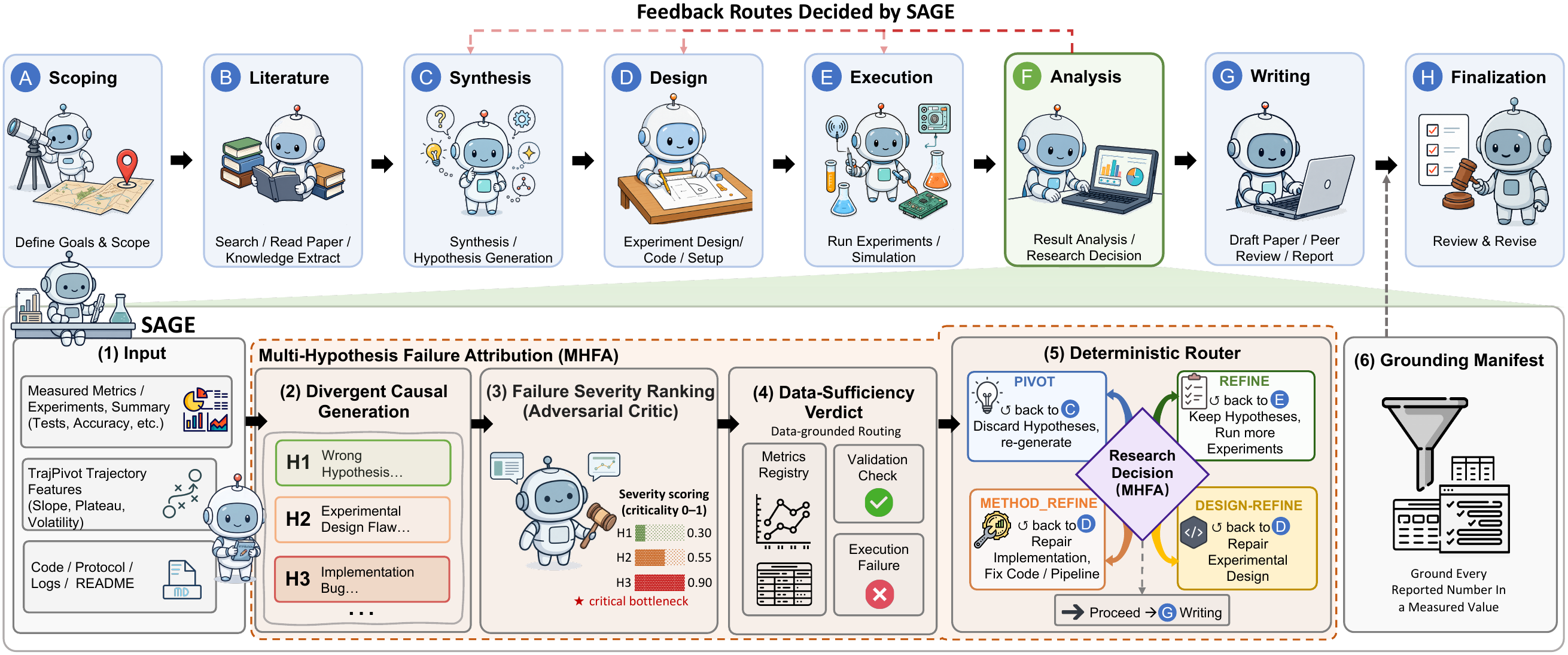}
    \caption{\textbf{Overview of \textsc{SAGE}.} The host pipeline runs eight phases, A (scoping) to H (finalization), shown on top. At the Analysis phase~(F), \textsc{SAGE} first detects zero-metric execution failures and routes them to an explicit repair path. For structural failures, \textbf{Multi-Hypothesis Failure Attribution} uses measured metrics, \textit{TrajPivot} trajectory features, and executed code/log summaries~(1) to generate candidate failure explanations~(2), rank them by severity~(3), and check data sufficiency~(4). A deterministic router~(5) maps the top verified structural cause to a hierarchy-aware repair, such as \textsc{Method-Refine}, \textsc{Design-Refine}, or \textsc{Pivot}, or proceeds to writing~(G). Finally, a grounding manifest~(6) check numeric table entries against measured values. The verdict clamp prevents hypothesis-discarding routes on valid but weak results, preserving honest negative findings.}
    \label{fig:overview}
\end{figure}

As illustrated in \cref{fig:overview}, \textsc{SAGE} executes this three-part recovery process through a systematic pipeline.
We first formally define the hierarchical recovery problem and the structured failure context (\cref{sec:formulation}). Operating over this context, our core module, \textbf{Multi-Hypothesis Failure Attribution (MHFA)}, executes a rigorous diagnostic process (\cref{sec:mhfa}). A deterministic router then maps the verified cause to the exact intervention level, safeguarded by data-sufficiency and regeneration bounds (\cref{sec:routing}). Finally, the pipeline concludes with a two-stage grounded reporting mechanism, guaranteeing that the drafted paper claims only what the executed pipeline actually measured (\cref{sec:reporting}).

\subsection{Problem Formulation and Context Representation}
\label{sec:formulation}
To formalize the recovery pipeline, we model recovery over a hierarchy of abstraction levels $\mathcal{L}=\{l_H, l_D, l_I\}$ representing hypothesis, design, and implementation, respectively. A completed experiment yields a terminal observation $\mathcal{O}_{end}$, such as a performance plateau far below the target or no usable metric. To recover, the system must identify whether intervention should occur at the hypothesis level, the experimental-design level, or the implementation level.
Monolithic reflection draws both cause and action from a single LLM sample; committing to the first sampled cause therefore dictates the action and collapses the posterior over causes to a single point. \textsc{SAGE} separates these steps. At analysis time, runs that emit no measured metric are handled by an explicit implementation-repair path. Remaining structural failures are passed to MHFA, which (i) samples a candidate set $\mathcal{H}=\{h_i\}_{i=1}^{K}$ of attributions; (ii) selects the most critical hypothesis $h^{*}$, with rectified category $c^{*}$ and score $s^{*}$; and (iii) emits an intervention $(A, \mathcal{L}_{target})=\pi(c^{*}, s^{*})$. The first two stages are stochastic LLM calls, while the third is a fixed symbolic rule, decoupling causal attribution from the execution action.

\paragraph{Structured Failure Context.} Before MHFA can generate hypotheses, it requires diagnostic context beyond flat logs. We construct a structured failure context $\mathcal{C}_{fail}=\langle \mathcal{T}, \mathcal{K} \rangle$. The knowledge component $\mathcal{K}$ stores the experiment's semantic stack: the current hypothesis, design configuration, result summary, and a structured summary of executed code and diagnosed deficiencies.

The trajectory component $\mathcal{T}$ is produced by \textit{TrajPivot}, an advisory stagnation detector. To ensure consistent interpretation across diverse scientific tasks, we direction-normalize all primary metrics $\mathcal{M}_t$ such that higher is always better (e.g., by negating error-based objectives like MASE or loss). Given an initial baseline measurement $\mathcal{M}_0$ and refinement iterations $t \in [1,T]$, the per-iteration gain is $g_t=\mathcal{M}_t-\mathcal{M}_{t-1}$. Using the first and last positive gains $g_{\mathrm{first}}, g_{\mathrm{last}}$, alongside the iteration and metric means $\bar{t}, \bar{\mathcal{M}}$, we summarize the trajectory via a marginal decay $\mathcal{D}$ and a regression slope $\mathcal{S}$:
\begin{equation}
    \mathcal{D} = 1 - \frac{g_{\mathrm{last}}}{\max(\epsilon, g_{\mathrm{first}})}, \qquad
    \mathcal{S} = \frac{\sum_{t=1}^{T}(t - \bar{t})\,(\mathcal{M}_t - \bar{\mathcal{M}})}{\sum_{t=1}^{T}(t - \bar{t})^2}.
\end{equation}
Here, $\epsilon$ is a small positive constant to prevent division by zero; if an experiment yields no positive gain, $\mathcal{D}$ defaults to $1.0$. Together with inter-run volatility and the gap to target, these features provide formal signatures for different failure modes: a decay near $1.0$ suggests stalled improvement, a near-zero slope with low volatility suggests saturation, and high volatility suggests instability rather than a fundamentally flawed hypothesis. Grounding attribution in $\mathcal{T}$ prevents the LLM from proposing repairs disconnected from the observed execution dynamics.

\subsection{Multi-Hypothesis Failure Attribution (MHFA)}
\label{sec:mhfa}
With the failure context $\mathcal{C}_{fail}$ defined, \textsc{SAGE} executes MHFA through a two-stage divergent-convergent diagnostic process.

\paragraph{Divergent Causal Generation.} Given $\mathcal{C}_{fail}$, the first stage of MHFA separates generation from evaluation. An agent adopting a ``senior-researcher'' persona is sampled to propose $K\in[3,5]$ diverse candidate attributions. Each candidate $h_i=\langle d_i, c_i, e_i, p_i \rangle$ contains a causal description $d_i$, a category $c_i \in \{\textsc{Method}, \textsc{Design}\}$, supporting evidence $e_i \subset \mathcal{C}_{fail}$, and a proposed structural fix $p_i$.
This two-category space is deliberate. \textsc{SAGE} treats low-level implementation faults, especially zero-metric failures, through a separate execution-repair path. Divergent attribution therefore focuses on structural deficits that local repair cannot resolve: an unsound core method (\textsc{Method}) or a misaligned experimental protocol (\textsc{Design}). If the candidate set $\mathcal{H}$ lacks categorical diversity, or if any $h_i$ fails to cite concrete evidence from $\mathcal{C}_{fail}$, generation is retried (\cref{app:impl}).

\paragraph{Failure Severity Ranking.} Scoring hypotheses with the same LLM instance that generated them invites self-endorsement. Therefore, the second stage uses an independent critic with a ``skeptical'' persona. The critic audits the logical soundness of $d_i$ against the cited evidence $e_i$, rectifies the category $c_i$ if needed, and assigns a criticality score $s_i \in [0,1]$ based on a fixed severity rubric, ranging from acceptable residual gaps to fatal directional failures (\cref{app:routing}). The top-scoring hypothesis $h^{*}=\argmax_i s_i$ becomes the intervention target, carrying its rectified category $c^{*}$ and verified score $s^{*}$.

\subsection{Deterministic Routing and Recovery Safeguards}
\label{sec:routing}

With a verified root cause $h^*$, the system must decide how to intervene. Allowing the LLM to act directly on its own diagnosis often causes semantic drift. To address this, we delegate the intervention to a deterministic routing framework, supported by two critical safeguards.

\paragraph{Hierarchy-Aware Deterministic Routing.} We apply a deterministic rule $\pi(c^{*}, s^{*})$ governed by two principles:
\begin{itemize}
    \item \emph{Severity gates depth:} The criticality score $s^{*}$ bounds the invasiveness of the intervention, ranging from accepting the result, to local refinement, to structural refinement, and finally to a pivot only for verified fatal flaws.
    \item \emph{Category and severity select the level:} A \textsc{Design} cause routes to the protocol level $l_D$. A \textsc{Method} cause routes to method refinement, preserving the hypothesis when the flaw is repairable and escalating to the hypothesis level $l_H$ only when the method failure is fatal.
\end{itemize}
This rule yields actions such as \textsc{Proceed}, \textsc{Local-Refine}, \textsc{Method-Refine}, \textsc{Design-Refine}, \textsc{Pivot}, and \textsc{Design-Pivot}, with the complete lookup specified in \cref{app:routing}. After a strict budget of \textsc{MaxPivots} (two in our runs), the pipeline proceeds with the best available measured outcome or abstains if no metric-bearing run exists. Because $\pi$ is deterministic, the same rectified diagnosis yields the same routing decision, making the recovery auditable.

\paragraph{Data-Sufficiency Grounding (The Verdict Clamp).} A purely criticality-driven router is blind to the distinction between a run that failed to produce usable evidence and a run that completed but returned a legitimately small or negative effect. A naive criticality-only $\pi$ can score both as severe and emit a hypothesis-discarding \textsc{Pivot}. This risks false hypothesis rejection on valid-but-weak results, burning the limited pivot budget and discarding reportable negative findings.

MHFA prevents this by grounding routing decisions in observed data. An independent LLM judge inspects the measured outcomes to return a verdict $v \in \{\textsc{Valid}, \textsc{ExecFail}\}$, a confidence $\kappa \in [0,1]$, and a short rationale. We then clamp the routed action $A=\pi(c^{*}, s^{*})$ using a confidence floor $\kappa_0=0.6$, where $\mathcal{P}$ is the set of hypothesis-discarding actions:
\begin{equation}
    A' = \Gamma(A, v, \kappa) =
    \begin{cases}
        \delta(A), & v = \textsc{Valid} \;\wedge\; \kappa \geq \kappa_{0} \;\wedge\; A \in \mathcal{P},\\
        A, & \text{otherwise.}
    \end{cases}
\end{equation}
Here, $\delta$ downgrades discarding actions to their non-discarding refinement equivalents, for example from \textsc{Pivot} to \textsc{Method-Refine}. The clamp fires only on a confident \textsc{Valid} verdict, so real but weak results are refined in place and steered toward an honest technical report rather than being discarded.

\paragraph{Failure-Aware Regeneration.} Once a routing decision triggers a rollback, such as a \textsc{Pivot}, the agent must avoid regenerating a near-identical hypothesis. Upon a pivot, MHFA archives a failure profile $\mathcal{A}_{fail}$ containing the failed hypothesis, trajectory features, and ineffective directions. This profile serves as a negative-knowledge prior. During regeneration, a distinctness gate compares each new candidate against $\mathcal{A}_{fail}$ and labels it as distinct, variant, or ambiguous. Non-distinct candidates are regenerated up to a retry budget. This mechanism steers generation away from failed directions, for example by encouraging protocol overhaul rather than a minor perturbation after a metric-insensitive failure.

\subsection{Grounded Reporting}
\label{sec:reporting}
\textsc{SAGE}'s reporting mechanism targets a narrower problem than full paper truthfulness: it constrains numeric result tables to values supported by measured artifacts. When asked to draft a results table, an autonomous LLM may introduce unmeasured means, standard deviations, baselines, or condition names. \textsc{SAGE} mitigates this numeric-table fabrication through a two-stage grounded reporting mechanism:

\textbf{1. Proactive Grounding Manifest:} Before drafting, \textsc{SAGE} derives a whitelist of permissible tabular data from the best experiment summary. This whitelist specifies the metrics, per-condition numbers, and pairwise contrasts that may be reported. Injected into the drafting prompt, it instructs the drafter to drop ``$\pm\text{std}$'' for metrics lacking variance measurements and to reserve unmeasured aggregates for qualitative discussion only.

\textbf{2. Reactive Arm-Agnostic Sanitizer:} Because prompt-time constraints are not sufficient on their own, a hard sanitizer acts as a backstop. Post-drafting, it checks every numeric table cell against a registry $\mathcal{R}$ of empirically measured values, allowing a $1\%$ relative tolerance for rounding. Any unverified value is replaced with a redaction sentinel (\texttt{-{}-{}-}). A verifier additionally refuses drafts that contain no measured metrics or fabricated condition names.

\emph{This method is scoped to reported numbers and table conditions.} Neither layer automatically verifies that the methods, datasets, libraries, model architectures, or statistical tests named in prose were actually instantiated in the executed code. We observe this \emph{method-provenance gap} in both \textsc{SAGE} and strong external baselines, and treat it as a central open problem for future autonomous scientific agents (\cref{sec:experiments}).

\section{Experiments}
\label{sec:experiments}

Our evaluation has two goals. First, we compare \textsc{SAGE} with a host-pipeline reflection baseline and with AI-Scientist-v2 under a blind, uniform artifact-level rubric (\cref{sec:exp-arcbench}). Second, we analyze what the recovery and reporting mechanisms actually contribute: which failures are recovered (\cref{sec:exp-recovery}), how the generated papers score under a strict main-conference-style bar and a calibrated autonomous-research bar (\cref{sec:exp-quality}), how the reporting gate behaves in practice (\cref{sec:exp-integrity}; with a worked case in \cref{app:b07}), and what can be inferred from the report-grounding analysis (\cref{sec:exp-ablation}).

A central observation guides the analysis. The generated papers still fall below a strict main-conference-style review bar, even when \textsc{SAGE} recovers executable experiments and real metrics. At the same time, a calibrated autonomous-research rubric separates systems by whether they produce complete, grounded, and honest artifacts. We report both views because their disagreement is informative: structured recovery improves execution and measurement, while final scientific writing and method-provenance grounding remain limiting factors.

\subsection{Setup}
\label{sec:exp-setup}
We evaluate \textsc{SAGE}, the self-correcting full autonomous research system of \cref{sec:method}. Its core contributions are a structured failure-recovery mechanism (MHFA with hierarchy-aware recovery and an execution-failure repair path) and a grounded numeric-reporting layer. For the stages orthogonal to these, namely scoping, literature search, experiment scaffolding, and writing, \textsc{SAGE} runs on a standard autonomous-research harness, \textsc{AutoResearchClaw}.

We compare against two systems. The first is our reflection baseline, \textsc{SAGE} w/o MHFA: the same harness with MHFA and hierarchy-aware routing disabled, so that failed runs fall back to free-form reflection. This is an ablation that holds the surrounding workflow fixed and isolates the structured recovery mechanism; we refer to it as \textsc{SAGE} w/o MHFA, or equivalently the baseline. The second is AI-Scientist-v2~\citep{aiscientistv2}, included as a strong external autonomous-scientist reference in \cref{sec:exp-arcbench}. To ensure a fair comparison, AI-Scientist-v2 artifacts are evaluated exactly as produced by that system, without applying \textsc{SAGE}'s reporting sanitizer or any other post-processing. The remaining diagnostic sections compare \textsc{SAGE} primarily against \textsc{SAGE} w/o MHFA, where the recovery mechanism is the principal intervention.

We run all systems in the same experimental campaign on a frozen 12-topic subset of \textsc{ARC-Bench}~\citep{autoresearchclaw}, an external open-ended autonomous research benchmark. The subset spans five domains: six topics in machine learning, two in statistics, two in quantum computing, one in biology, and one in high-energy physics; the selection protocol and full topic list are given in \cref{app:topic-selection}. For a controlled comparison, all systems, including AI-Scientist-v2, are rerun on the identical topic set using the same backbone model and model snapshot, API configuration, per-topic LLM and experiment-execution budget caps, wall-clock limit, hardware resources, and software environment. We define a run as \emph{metrics-bearing} if its experiment stage emits at least one task-relevant measured metric, rather than only run counts, logs, or timing information. We re-draft each \textsc{SAGE} paper once with report-grounding enabled, which can change the written artifact but not the underlying measurements; no \textsc{SAGE}-specific sanitizer or post-processing is applied to AI-Scientist-v2 artifacts. Each paper is assessed under two complementary bars: a \emph{strict main-conference-style bar}, where a blind reviewer (Claude Opus 4.8, using a NeurIPS-style form on paper text only) scores it out of ten, and a \emph{calibrated autonomous-research bar}, AR-Eval (\cref{app:areval}), which credits honest abstention and penalizes genuine grounding or integrity failures. For transparency, a complete end-to-end \textsc{SAGE} run costs approximately $\$10$ to $\$20$ in LLM API calls, varying mainly with the number of recovery iterations.

\subsection{Comparison to Other Autonomous Scientists}
\label{sec:exp-arcbench}
We compare \textsc{SAGE} with AI-Scientist-v2~\citep{aiscientistv2} and \textsc{AutoResearchClaw} (the host reflection baseline) across the twelve topics described in \cref{sec:exp-setup}. Each complete submission, including experiment code, machine-readable results, and the generated paper, is scored on a 0 to 100 scale by a blind reviewer under a uniform rubric. As shown in \cref{tab:arcbench}, \textsc{SAGE} obtains the highest overall score in this evaluation (52.0), compared with AI-Scientist-v2 (48.2) and the host reflection baseline (24.8). Its advantage is concentrated in code development and execution, while result-analysis quality remains low.

\paragraph{Protocol.}
Each submission is scored by an independent blind reviewer (Claude Opus 4.8) that reads only that submission, with framework identity stripped from file paths. The rubric evaluates three dimensions, aggregated in a $2{:}2{:}3$ ratio: \emph{Code Development} (whether the code is a genuine implementation rather than a stub), \emph{Code Execution} (whether it runs and produces real metrics with genuine replication), and \emph{Result Analysis} (whether the written quantitative and methodological content is grounded in the code and results). A topic where an arm produces no paper scores zero. Thus, all means are completion-weighted over the twelve topics. We use the same auditing protocol across arms: the judge receives the full code project, the results artifact matched to the dimension being evaluated, and the generated paper.

\begin{table}[t]
\renewcommand{\arraystretch}{1.0}
\centering\small
\caption{Blind, per-submission, uniform artifact-level evaluation across twelve topics. Each dimension is in $[0,100]$; \emph{Overall} is the $2{:}2{:}3$ weighted mean. \emph{Wins} counts per-topic head-to-head victories against AI-Scientist-v2 where both systems produced comparable artifacts. \textsc{SAGE} leads overall and in both code dimensions, while result analysis remains low for both autonomous-scientist systems and differs by less than the judge noise.}
\label{tab:arcbench}
\begin{tabular}{lccccc}
\toprule
Framework & Code Dev. & Code Exec. & Result Analysis & \textbf{Overall} & Wins\\
\midrule
AI-Scientist-v2 & 58.3 & 51.7 & \textbf{39.2} & 48.2 & 2\\
\textsc{AutoResearchClaw} & 33.3 & 20.8 & 21.7 & 24.8 & -\\
\textsc{SAGE} (ours) & \textbf{67.5} & \textbf{62.5} & 34.6 & \textbf{52.0} & \textbf{7}\\
\bottomrule
\end{tabular}
\end{table}

\paragraph{Where the gains lie.}
\textsc{SAGE}'s gains are concentrated in the two code-oriented dimensions: Code Development (67.5 vs.\ 58.3 vs.\ 33.3) and Code Execution (62.5 vs.\ 51.7 vs.\ 20.8). This matches the intended effect of failure recovery: the system more often repairs experiments into runnable, metrics-bearing artifacts. These scores should not be read as evidence that \textsc{SAGE} solves autonomous scientific writing. Result Analysis remains weak, and the gap between \textsc{SAGE} and AI-Scientist-v2 on that dimension is within single-judge noise.

\paragraph{A method-provenance gap.}
The low Result Analysis scores expose a broader limitation. For \textsc{SAGE}, headline numbers generally trace back to measured runs, such as ML20 sMAPE 15.75 and B07 growth 0.4678. However, the prose can still misstate the methodology: generated Method sections sometimes name libraries, models, datasets, or statistical tests that the executed code did not use. Similar issues appear in AI-Scientist-v2 artifacts, such as claiming a MadGraph and Pythia pipeline while the executed code fits a toy analytic form to self-generated labels, or claiming Qiskit with a noise model while the code uses PennyLane without such a model. In \textsc{SAGE}, one artifact names the Friedman-1 dataset when the executed run used California Housing. The reporting gate grounds numeric table cells, but it does not verify prose-level method claims. We call this open problem \emph{method-provenance grounding}: verifying that methodological claims in the paper are backed by concrete executed artifacts.

\paragraph{Case studies.}
\Cref{tab:cases} highlights three topics where the blind judge preferred \textsc{SAGE} over AI-Scientist-v2. The cases are not meant to prove broad scientific superiority; they illustrate the kinds of artifact-level differences captured by the rubric.

\begin{table}[t]
\centering\footnotesize
\renewcommand{\arraystretch}{1.0}
\setlength{\tabcolsep}{2pt}
\caption{\textbf{Qualitative Case Studies.} Three representative topics from the head-to-head comparison (\cref{sec:exp-arcbench}) where the blind judge preferred \textsc{SAGE} over AI-Scientist-v2. In each instance, \textsc{SAGE} conducts original, verifiable, and structurally deep scientific exploration, whereas the baseline tends to reproduce standard scikit-learn demonstrations or veer off-topic.}
\label{tab:cases}
\begin{tabular}{@{}l l p{6.5cm} p{4.5cm}@{}}
\toprule
Topic & Strength & \textsc{SAGE} & AI-Scientist-v2\\
\midrule
ML12 (clustering) & Methodological depth & Crossed-factorial design isolating the algorithm-selector confound; decomposes the gap into ${\sim}1/4$ selector and ${\sim}3/4$ geometry & Reproduces the standard scikit-learn demonstration on one small dataset\\
\addlinespace
Q01 (quantum) & Verifiable finding & Finds IQP $\equiv$ ZZ implementation-equivalence, confirmed by a unitary check to machine precision & Reports the expected result that quantum trails a classical baseline\\
\addlinespace
S01 (bootstrap CI) & Honest recovery & Detects a silent harness failure, refuses an unsupported ranking, and ships a reusable audit protocol & Off-topic: benchmarks regression ensembles, not bootstrap intervals\\
\bottomrule
\end{tabular}
\end{table}

\subsection{Failure Recovery}
\label{sec:exp-recovery}
\paragraph{Outcome-level recovery.}
\textsc{SAGE} yields metrics-bearing results on 11 of 12 topics, compared with 5 of 12 for \textsc{SAGE} w/o MHFA. The dominant pathology in the failing runs is a non-runnable or non-measuring experiment: a \texttt{main.py} that is only a \texttt{Config} dataclass, a leaked prose preamble that never executes, code that omits the stated models, or a harness without an experiment loop. At analysis time, \textsc{SAGE}'s router judges what failed and at which level, then routes accordingly: zero-metric execution failures to an explicit repair path, and structural failures to MHFA attribution and hierarchy-aware recovery. The higher completion follows from this structured routing as a whole, both the execution repair and the structural re-diagnosis that escalates beyond it, rather than from any single repair heuristic. \Cref{tab:mhfa-mechanism} gives the per-topic breakdown, including which of the baseline's shortfalls are method-attributable and which two (P03, S01) are infrastructure confounds excluded from the comparison.

\paragraph{Recovery quality on hard cases.}
We next illustrate the range of \textsc{SAGE}'s recovery outcomes on four hard cases that span the clean, flagged, and failed gates: one clean recovery (P03), two flagged recoveries (ML20, S01), and one bounded non-recovery (S02). P03 is the clean case: after a flat metric with zero slope, \textsc{SAGE} repeatedly routed recovery to \textsc{Method-Refine} and obtained a metric-bearing run with a clean gate. In the flagged cases, \textsc{SAGE} recovers real metrics but the artifact retains a defect. For ML20, after an entry point that produced no experiment, \textsc{SAGE} diagnosed a \texttt{zero\_metrics\_code\_issue}, repaired the entry point, re-ran the experiment, and then applied a downstream \textsc{Method-Refine} that corrected the MASE denominator, yielding sMAPE, MASE, and coverage metrics across 5 of 5 forecasters, yet the final artifact still carried reporting or quality flags. S01 similarly recovered real metrics but expanded to an excessive 216 conditions. S02 is the bounded non-recovery: after the recovery budget was exhausted, \textsc{SAGE} produced no metric-bearing run and abstained from writing a paper, whereas \textsc{SAGE} w/o MHFA produced a degraded artifact with unsupported claims. This bounded behavior does not turn every failed run into a paper; it prevents unbounded repair loops and records an honest non-result.

\paragraph{Per-topic self-correction versus \textsc{SAGE} w/o MHFA.}
\Cref{tab:mhfa-mechanism} traces \textsc{SAGE}'s recovery trajectories across all twelve topics, paired with \textsc{SAGE} w/o MHFA, i.e.\ the same host pipeline with MHFA disabled and recovery reverting to free-form reflection. The central contrast is structural and holds independently of any run outcome. Across all twelve topics, the baseline's recovery vocabulary is confined to a single action (\textsc{Refine}, emitted at most twice) and never escalates to the method, design, or hypothesis levels. In stark contrast, \textsc{SAGE} executes a fresh divergent-then-critic pass on every retry and, in 7 of the 11 recovered topics, successfully escalates across abstraction levels as evidence accumulates. For example, once an entry-point repair unblocks a zero-metric run, MHFA re-diagnoses ML20 as a seasonal-naive \textsc{MASE}-denominator error and S01 as bit-identical condition degeneracy. Both are deep structural causes that a single \textsc{Refine} loop cannot reach. The most severe hypothesis-level \textsc{Pivot} fires on only two topics (ML01, ML02). On the remaining high-severity \textsc{Method} causes, the data-sufficiency clamp explicitly downgrades the \textsc{Pivot} to \textsc{Method-Refine}, successfully trading aggressive hypothesis rejection for the in-place refinement of valid-but-weak results.

The table separates the results into two regimes. In the top block (directly comparable topics where the baseline produced a deliverable), the calibrated AR-Eval standard awards \textsc{SAGE} a win on three topics, a tie on one, and favors the baseline on two. We transparently analyze rather than hide these two reversals: B07 is a writer-to-sanitizer coordination failure (\cref{app:b07}), and S02 represents an honest abstention by \textsc{SAGE} due to a lack of metric-bearing runs, whereas the baseline hallucinates a degraded paper with unsupported claims. The bottom block lists topics where the baseline produced no deliverable. Crucially, on four of these (ML20, ML12, Q01, Q03), the absence is a genuine method-attributable failure. The baseline hits a zero-metric or degenerate-code condition. For instance, generating a \texttt{main.py} that leaks an LLM prose preamble and raises a \texttt{SyntaxError}, executing an experiment that exits in 0.03\,s with empty metrics, or writing code that entirely omits the stated models. After a single \textsc{Refine} attempt, it fails to recover, and the writer then honestly declines to draft a paper without real data. \textsc{SAGE}, however, recovers all four through repeated re-diagnosis, yielding a clean artifact gate on three (ML12, Q01, Q03). Only P03 (a harness \texttt{TypeError} during code generation) and S01 (a transient model-call failure) are true non-method confounds; we rigorously exclude these from the comparison rather than unjustly charge them to the baseline.

\begin{table*}[t]
\centering
\footnotesize
\renewcommand{\arraystretch}{1.1}
\caption{\textbf{Per-Topic Self-Correction Traces.} \textsc{SAGE} successfully recovers 11/12 topics by dynamically escalating its repairs across abstraction levels (from \textsc{Local-Refine} to \textsc{Method-Refine} or \textsc{Pivot}). In stark contrast, the \textsc{SAGE} w/o MHFA baseline is trapped in a single \textsc{Refine} action and fails completely on 6 topics. Notably, \textsc{SAGE} strictly limits its budget to avoid endless loops (honestly abstaining on S02 rather than fabricating) and successfully rescues four deep method-attributable failures (ML20, ML12, Q01, Q03) that caused the baseline to fail outright.}
\label{tab:mhfa-mechanism}
\begin{tabular}{@{}l
>{\raggedright\arraybackslash}p{0.45\textwidth}
>{\raggedright\arraybackslash}p{0.25\textwidth}
c c@{}}
\toprule
\textbf{Topic} & \textbf{\textsc{SAGE}: Re-diagnosis chain (cause $\Rightarrow$ action)} & \textbf{\textsc{SAGE} w/o MHFA} & \textbf{AR-Eval} & \textbf{Status} \\
\midrule
\multicolumn{5}{@{}l}{\textit{Directly comparable: baseline produced a deliverable}}\\
\midrule
ML01 & 
1. no \texttt{main()} entry $\Rightarrow$ \textsc{Local-Refine}\newline
2. phantom run $\Rightarrow$ \textsc{Pivot}\newline
3. silent arm aggregation $\Rightarrow$ \textsc{Method-Refine} & 
\textsc{Refine} $\times 2$, no escalation & \textbf{8} / 6 & \textsc{Flagged} \\
\addlinespace
ML02 & 
1. phantom run, $K{=}0$ no usable data $\Rightarrow$ \textsc{Pivot}\newline
2. (hypothesis regenerated) $\Rightarrow$ \textsc{Pivot} & 
\textsc{Refine} $\times 2$, no escalation & \textbf{7} / 5 & \textsc{Flagged} \\
\addlinespace
ML18 & 
1. no \texttt{main()} entry $\Rightarrow$ \textsc{Local-Refine}\newline
2. runtime failure in \texttt{run\_one} $\Rightarrow$ \textsc{Design-Refine}\newline
3. flat metric / agg.\ bug $\Rightarrow$ \textsc{Method-Refine} & 
\textsc{Refine} $\times 2$, no escalation & \textbf{8} / 5 & \textsc{Flagged} \\
\addlinespace
ML16 & 
1. immediate plateau ($199.3\!\to\!182.8$) $\Rightarrow$ \textsc{Method-Refine}\newline
2. regret metric mis-targeted $\Rightarrow$ \textsc{Design-Refine} & 
\textsc{Refine} $\times 2$, no escalation & 4 / 4 & \textsc{Flagged} \\
\addlinespace
B07  & 
1. metric collapse $0.468\!\to\!0.05$; non-discriminative\newline
2. still flat $\Rightarrow$ \textsc{Design-Refine} $\times 3$ (latent defect survives) & 
\textsc{Refine} $\times 2$ & 5 / \textbf{6} & \textsc{Flagged} \\
\addlinespace
S02  & 
1. no \texttt{main()} entry $\Rightarrow$ \textsc{Local-Refine} $\to$ \textsc{Method-Refine}\newline
2. budget exhausted $\Rightarrow$ \emph{writer abstains} & 
\textsc{Refine} $\times 2$, degraded paper & abstain / \textbf{4} & \textsc{Fail} \\
\midrule
\multicolumn{5}{@{}l}{\textit{SAGE-only: baseline produced no deliverable (cause of absence noted)}}\\
\midrule
ML20 & 
1. no \texttt{main()} entry $\Rightarrow$ \textsc{Local-Refine}\newline
2. \textsc{MASE} $23/49$ implausible $\Rightarrow$ \textsc{Method-Refine} & 
method: codegen omits the models & \textbf{6} / n/a & \textsc{Flagged} \\
\addlinespace
ML12 & 
1. empty \texttt{config\_snapshot} $\Rightarrow$ \textsc{Method-Refine}\newline
2. flat metric $0.8447$ $\Rightarrow$ \textsc{Design-Refine}\newline
3. monotonic non-improvement $\Rightarrow$ \textsc{Method-Refine} & 
method: prints config, no metrics & \textbf{7} / n/a & \textsc{Clean} \\
\addlinespace
Q01  & 
1. metric saturated at $0.9784$ $\Rightarrow$ \textsc{Design-Refine} $\times 2$\newline
2. still flat $\Rightarrow$ \textsc{Method-Refine} & 
method: leaked preamble, \texttt{SyntaxError} & \textbf{7} / n/a & \textsc{Clean} \\
\addlinespace
Q03  & 
flat sentinel $2.11{\times}10^{8}$; zero slope $\Rightarrow$ \textsc{Method-Refine} $\times 3$ & 
method: 0.03s run, empty metrics & \textbf{7} / n/a & \textsc{Clean} \\
\addlinespace
P03  & 
1. flat metric; collapse $+1.02\!\to\!-0.43$\newline
2. oscillation $\Rightarrow$ \textsc{Method-Refine} $\times 3$ & 
confound: \texttt{TypeError} (codegen) & \textbf{6} / n/a & \textsc{Clean} \\
\addlinespace
S01  & 
1. no \texttt{main()} entry $\Rightarrow$ \textsc{Local-Refine}\newline
2. conditions not varied $\Rightarrow$ \textsc{Method-Refine}\newline
3. bit-identical outputs $\Rightarrow$ \textsc{Method-Refine} & 
confound: model-call failure & \textbf{7} / n/a & \textsc{Flagged} \\
\bottomrule
\end{tabular}
\end{table*}

\subsection{Dual-Standard Evaluation of Artifact Quality}
\label{sec:exp-quality}
Recovering execution metrics is necessary but not sufficient. We therefore evaluate generated papers under two standards. The strict main-conference-style bar asks whether the paper would be acceptable as a conventional venue submission. AR-Eval asks a different question: whether an autonomous pipeline produced a complete, empirically grounded, and honest artifact. The disagreement between the two bars is part of the finding.

\paragraph{Strict main-conference-style standards.}
Under the strict paper-only venue bar, generated papers from both arms fall below the acceptance-level quality. In the round evaluating non-grounded \textsc{SAGE} drafts and baseline papers, all 12 papers score between 2 and 3 out of 10, collapsing the arm-level gap to roughly 0.3 points. A fresh blind round over the 8 report-grounded \textsc{SAGE} deliverables averages 2.25/10, with all 8 still showing blank or degraded result tables. The main defects are prose numbers that contradict blanked table cells and experimental designs that are too narrow, often single-seed or single-cell.

\begin{figure}[t]
\centering
\begin{minipage}[c]{0.62\linewidth}\centering
\includegraphics[width=\linewidth]{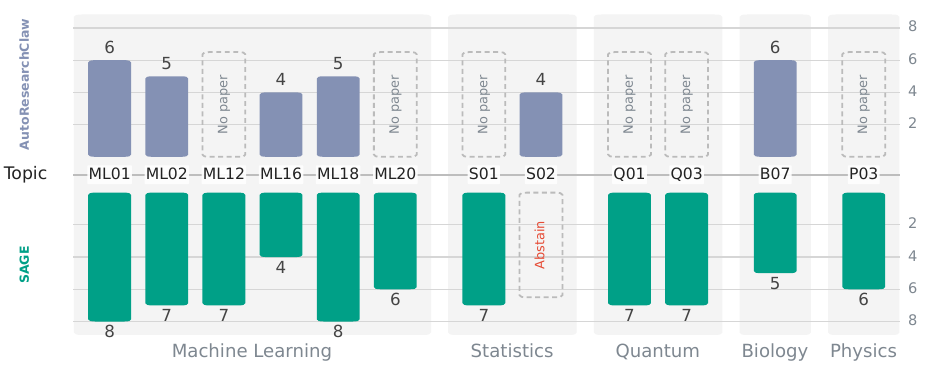}
\end{minipage}\hfill
\begin{minipage}[c]{0.35\linewidth}\centering
\setlength{\tabcolsep}{4pt}
\resizebox{\linewidth}{!}{%
\begin{tabular}{l *{6}{c} @{\hspace{1.1em}} *{6}{c}}
\toprule
 & \multicolumn{6}{c}{\textsc{SAGE}} & \multicolumn{6}{c}{\textsc{SAGE} w/o MHFA}\\
\cmidrule(lr){2-7}\cmidrule(l){8-13}
Topic & AC & EG & RI & MS & SC & \textbf{/10} & AC & EG & RI & MS & SC & \textbf{/10}\\
\midrule
ML01 & 5 & 4 & 3 & 4 & 4 & \textbf{8} & 4 & 2 & 4 & 3 & 3 & \textbf{6}\\
ML02 & 4 & 4 & 3 & 4 & 4 & \textbf{7} & 3 & 2 & 3 & 3 & 3 & \textbf{5}\\
ML12 & 4 & 4 & 3 & 4 & 4 & \textbf{7} & - & - & - & - & - & -\\
ML16 & 3 & 2 & 2 & 3 & 3 & \textbf{4} & 2 & 2 & 2 & 3 & 3 & \textbf{4}\\
ML18 & 5 & 4 & 4 & 4 & 4 & \textbf{8} & 4 & 3 & 2 & 3 & 3 & \textbf{5}\\
ML20 & 4 & 3 & 2 & 4 & 4 & \textbf{6} & - & - & - & - & - & -\\
\midrule
S01 & 3 & 4 & 4 & 3 & 4 & \textbf{7} & - & - & - & - & - & -\\
S02 & - & - & - & - & - & - & 3 & 2 & 2 & 3 & 3 & \textbf{4}\\
\midrule
Q01 & 3 & 4 & 3 & 3 & 4 & \textbf{7} & - & - & - & - & - & -\\
Q03 & 3 & 4 & 5 & 4 & 4 & \textbf{7} & - & - & - & - & - & -\\
\midrule
B07 & 3 & 3 & 2 & 3 & 3 & \textbf{5} & 4 & 2 & 5 & 4 & 3 & \textbf{6}\\
\midrule
P03 & 3 & 3 & 4 & 3 & 4 & \textbf{6} & - & - & - & - & - & -\\
\bottomrule
\end{tabular}}
\end{minipage}
\caption{AR-Eval results on the 12-topic subset, all from the primary judge (Claude Opus 4.8). Left: per-topic paper quality out of 10. Right: per-dimension sub-scores that aggregate into each score, with AC for autonomy and completion, EG for empirical grounding, RI for reporting integrity, MS for methodological soundness, and SC for scientific coherence. A dash marks topics where that arm produced no paper, or where \textsc{SAGE} abstained rather than report unverified metrics. \textsc{SAGE} averages 6.75/10 over its eight canonical deliverables and approximately 6.5/10 over all eleven grounded papers, compared with 5.00/10 for the six baseline deliverables. B07 is a baseline-favored reversal analyzed in \cref{app:b07}.}
\label{fig:quality}
\end{figure}

\paragraph{Calibrated AR-Eval standards.}
Under general AR-Eval, the narrative is different (\cref{fig:quality}). The eight canonical \textsc{SAGE} deliverables average 6.75/10 (2 strong, 5 credible, 1 partial), compared with 5.00/10 for the six baseline deliverables (0 strong, 2 credible, 4 partial). Evaluating all eleven grounded \textsc{SAGE} papers, including Q03, B07, and P03, yields approximately 6.5 against the same baseline mean. \textsc{SAGE} earns a credible artifact on six topics where the baseline produced no paper. On the five topics where both arms scored, \textsc{SAGE} exceeds the baseline on three, ties on one, and trails on B07. Two cases favor the baseline: B07, where a writer-sanitizer coordination failure lowers \textsc{SAGE}'s reporting-integrity score (\cref{app:b07}), and S02, where \textsc{SAGE} abstains due to the absence of metrics. These results support a narrow conclusion: \textsc{SAGE} produces more complete autonomous-research artifacts, but neither arm produces main-conference-quality papers.

\paragraph{Judge reliability.}
Because AR-Eval initially rests on a single judge, we run a reliability check with a second judge model. Claude Opus 4.8 and Claude Sonnet 4.6 blindly score all 14 generated papers, consisting of the 8 canonical grounded \textsc{SAGE} deliverables and the 6 baseline deliverables, under the same rubric. Both judges rank \textsc{SAGE} above the baseline on average: the gap is +1.75 under Opus and +0.92 under Sonnet. We nevertheless treat this as a diagnostic reliability check rather than a substitute for human expert validation. The sample size is small, and disagreement is concentrated in methodological soundness, where rank agreement is low. We therefore report AR-Eval alongside the strict main-conference-style review bar, not instead of it.

\paragraph{Human expert evaluation.}

\begin{wraptable}{r}{0.38\linewidth}
\centering\tiny
\setlength{\tabcolsep}{6pt}
\vspace{-\baselineskip}
\caption{Blind human evaluation by three ML-PhD reviewers on the six ML topics (Overall, 1--10; mean over topics and reviewers). \textsc{SAGE} leads on Overall (paired Wilcoxon $p=0.031$ vs.\ AI-Scientist-v2, $p=0.062$ vs.\ \textsc{AutoResearchClaw}, where $N{=}5$ as it produced no paper on ML20).}
\label{tab:human}
\resizebox{0.85\linewidth}{!}{%

\begin{tabular}{@{}l c@{}}
\toprule
System & Overall (1--10)\\
\midrule
\textsc{SAGE} (ours)        & \textbf{5.67}\\
AI-Scientist-v2             & 4.72\\
\textsc{AutoResearchClaw}   & 3.00\\
\bottomrule
\end{tabular}
}
\end{wraptable}
To test whether AR-Eval merely reflects an LLM judge rewarding an LLM-built artifact, we ran a blind human evaluation on the six machine-learning topics, the subset where our three reviewers (all ML PhDs) have domain expertise. Each reviewer scored three anonymized versions per topic, \textsc{SAGE}, \textsc{AutoResearchClaw}, and AI-Scientist-v2, on the same dimensions and blind to the system mapping. The human ranking matches the ordering from our automated evaluations (\cref{tab:human}): \textsc{SAGE} attains the highest overall score (5.67/10), above AI-Scientist-v2 (4.72) and well above \textsc{AutoResearchClaw} (3.00). Crucially, the per-topic human and AR-Eval overall scores for \textsc{SAGE} correlate at Spearman $\rho=0.61$, comparable to the $\rho=0.625$ between our two LLM judges; AR-Eval therefore agrees with domain experts about as well as two strong models agree with each other. The written comments echo the same picture: \textsc{SAGE} is consistently credited with controlled, falsifiable designs, explicit confound isolation, and clearly posed research questions, the main reservation being the limited experimental scale attainable in a single autonomous run. \textsc{AutoResearchClaw} is faulted more fundamentally for incomplete or unexecuted experiments and malformed, placeholder-filled result tables, and AI-Scientist-v2 for resting on narrow single-dataset evidence with overly broad claims despite clear writing.

\subsection{Reporting-Integrity Behavior}
\label{sec:exp-integrity}
\Cref{sec:exp-arcbench} exposed a method-provenance gap. The more visible numeric symptom is prose-table inconsistency: the writer may state a number in prose while the sanitizer blanks the corresponding table cell as unverified. This behavior occurs because the reporting-integrity gate fires symmetrically on both arms and redacts any table number it cannot trace to a measured value. It improves numeric caution but can make the final paper internally inconsistent when prose and tables are not coordinated.

This coordination defect caps scores under both evaluation bars. For example, the ML01 deliverable states a Concrete-dropout accuracy standard deviation of 0.0002 in prose, while the corresponding table cell is correctly blanked as unverified. The sanitizer is doing its job, but the writer has not been forced to revise the surrounding prose. This explains why reporting integrity often remains in the 2 to 3 range despite real measured metrics.
We also audited the gate itself. During evaluation, we identified and patched a thousands-separator parsing bug that incorrectly blanked real grounded values, for example turning $13{,}365{,}000$ into $13{,}\texttt{-{}-{}-}{,}000$. The Q03 deliverable originally had 24 such values incorrectly blanked, which are now restored. This was a parser bug rather than a failure of the grounding principle.

\subsection{Ablation: MHFA Components}
\label{sec:exp-ablation}
\begin{wraptable}{r}{0.35\linewidth}
\centering\tiny
\vspace{-\baselineskip}
\caption{Ablation of MHFA components on the 12-topic subset, scored by the AR-Eval judge (Claude Opus 4.8). Each row removes one component from the full system; AR-Eval is the range (min--max) over the 12 papers.}
\label{tab:ablation}
\resizebox{\linewidth}{!}{%
\begin{tabular}{@{}l c@{}}
\toprule
Configuration & AR-Eval Score\\
\midrule
\textsc{SAGE} (full)                     & 5--8\\
\quad w/o independent critic             & 5--7\\
\quad w/o divergent generation ($K{=}1$) & 3--7\\
\quad w/o MHFA                           & 0--6\\
\bottomrule
\end{tabular}
}
\end{wraptable}
The recovery gains above could come from the surrounding repair machinery rather than from the multi-hypothesis structure itself. To separate the two, we ablate three components of MHFA and score the resulting papers with the same AR-Eval judge (Claude Opus 4.8) used in \cref{sec:exp-quality}. We remove, in turn, divergent generation (leaving a single hypothesis, $K{=}1$), the independent critic (so the generator scores its own hypotheses), and MHFA in its entirety (leaving only the zero-metric code-repair path). Each ablation degrades quality in a distinct and interpretable way (\cref{tab:ablation}). With a single hypothesis, a misattributed cause is never corrected: the run commits to a surface-level fix and ships a thinner, degraded paper. Without the critic, self-endorsement bias lets the generator rank a secondary cause as most critical, so the router intervenes at the wrong level. Without MHFA, the system has no structured diagnosis at all and either writes on a broken experiment or blindly code-repairs, frequently degrading or failing outright. Only the full system routes the verified critical cause to the appropriate intervention and recovers clean, metric-bearing artifacts. This indicates that the multi-hypothesis structure, not the repair path alone, is what drives reliable recovery.

\section{Conclusion}
\label{sec:conclusion}
Autonomous AI scientists hold immense promise, yet their inability to robustly recover from experimental failures has constrained them to brittle, linear workflows. In this paper, we identified monolithic reflection as the root cause of this fragility and introduced \textsc{SAGE}, an autonomous agent that tackles failure through structured causal diagnosis. By implementing Multi-Hypothesis Failure Attribution (MHFA), \textsc{SAGE} successfully replaces blind trial-and-error with a rigorous cycle of divergent generation, independent severity ranking, and deterministic, hierarchy-aware routing. 
Coupled with a strict, two-stage grounded reporting mechanism, \textsc{SAGE} not only achieves state-of-the-art performance across highly diverse scientific domains (outperforming AI-Scientist-v2) but also radically improves the recovery rate of metrics-bearing experiments. More importantly, our commitment to extreme reporting transparency reveals a critical \emph{bottleneck shift} in the field. Once the diagnostic failure bottleneck is resolved by MHFA, the true limits of current autonomous research emerge: faithful implementation and the method-provenance gap. Ultimately, \textsc{SAGE} establishes a more trustworthy paradigm for AI scientists: one that systematically learns from its own mistakes, resolutely refuses to fabricate results, and reports only what it can empirically verify.

\bibliographystyle{plainnat}
\bibliography{references}

\newpage
\appendix

\section{Topic Selection Protocol}
\label{app:topic-selection}

To rigorously assess open-ended scientific discovery, we source our evaluation tasks from a comprehensive multi-domain benchmark~\citep{autoresearchclaw}. Specifically, we evaluate \textsc{SAGE} on a \textbf{predefined subset of 12 distinct research topics} spanning five major scientific domains: machine learning, high-energy physics, quantum computing, biology, and statistics (\cref{tab:topic-list}). 

\paragraph{Why evaluate on a 12-topic subset?} 
Our primary objective is not merely to exhaust a specific dataset, but to construct and rigorously validate a new, self-correcting paradigm for fully autonomous research. To achieve this systematically, we employ a ``core-plus-generalization'' evaluation strategy. First, we anchor our core evaluation in the machine learning (ML) domain (6 topics), utilizing it as the primary testbed to verify the fundamental efficacy of our failure recovery and reporting mechanisms. Second, we extend our evaluation to the remaining four domains (statistics, quantum computing, biology, and physics) to serve as a strict \emph{generalization validation}. This cross-domain testing \textbf{demonstrates} that \textsc{SAGE}'s structured diagnostic capabilities are not overfitted to ML engineering, but are genuinely effective across diverse scientific fields.

To guard against post-hoc selection bias and ensure the integrity of our evaluation, we strictly adhered to two rules during this curation process:
\begin{enumerate}
  \item \textbf{Frozen before execution.} The topic list was fixed prior to running any system; no topic was added or removed after observing results.\footnote{Authors' protocol statement: the subset was committed before the runs reported here.}
  \item \textbf{All attempted topics reported.} Every topic we attempted appears in the results, including \textsc{S02}, where \textsc{SAGE} exhausted its repair budget, produced no metric-bearing run, and \emph{honestly declined to write a paper}. We report this non-result rather than dropping the topic to inflate our success rate.
\end{enumerate}

Furthermore, we openly acknowledge that the high computational and LLM API resource costs required to run end-to-end, multi-iteration autonomous research pipelines inherently limit our current evaluation scale. We treat this resource constraint as a practical limitation of the field today. Nevertheless, establishing this 12-topic, cross-domain foundation is a crucial milestone. Moving forward, as inference costs decrease, we will continue to expand this evaluation scale and persistently build toward an even more reliable, trustworthy, and scalable paradigm for fully autonomous scientific discovery.

\begin{table}[h]
\centering
\small
\setlength{\tabcolsep}{5pt}
\renewcommand{\arraystretch}{1.15}
\caption{The 12 evaluated scientific research topics and their corresponding domains. Domain counts:
ML 6, statistics 2, quantum 2, physics 1, biology 1 (all five domains represented).}
\label{tab:topic-list}
\begin{tabular}{@{}l l l@{}}
\toprule
\textbf{Topic} & \textbf{Domain} & \textbf{Task (one line)} \\
\midrule
ML01 & ML         & Dropout variants for tabular MLPs          \\
ML02 & ML         & Ensemble regressors under noise            \\
ML12 & ML         & Clustering: algorithm vs.\ selector effect \\
ML16 & ML         & Bandit regret decomposition                \\
ML18 & ML         & Minority-aware post-hoc calibration        \\
ML20 & ML         & Classical seasonal forecasters             \\
P03  & Physics    & $B\!-\!L$ $Z'$ dilepton exclusion recast   \\
Q01  & Quantum    & VQC data-encoding audit                    \\
Q03  & Quantum    & VQE optimizers under shot noise            \\
B07  & Biology    & FBA variants on \textit{E.\ coli}          \\
S01  & Statistics & Bootstrap coverage audit                   \\
S02  & Statistics & Cross-fit / DML ATE estimation             \\
\bottomrule
\end{tabular}
\end{table}

\section{Routing Specification}
\label{app:routing}

\paragraph{Severity rubric.} The critic $V_\phi$ assigns criticality scores according to the following calibrated bands: $s_i \in [0, 0.4)$ denotes an acceptable residual gap (the result, while suboptimal, does not warrant structural intervention); $[0.4, 0.5)$ a locally correctable issue such as hyperparameter misconfiguration; $[0.5, 0.8)$ a major structural flaw in the method or experimental design; and $[0.8, 1.0]$ a fatal directional failure invalidating the current hypothesis or protocol.

\paragraph{Routing rule.} Table~\ref{tab:routing} and Algorithm~\ref{alg:routing} give the complete specification of $\pi(c^*, s^*)$.

\begin{table}[ht]
\centering
\caption{The deterministic routing rule $\pi(c^*, s^*)$.}
\label{tab:routing}
\small
\begin{tabular}{llll}
\toprule
Criticality $s^*$ & Category $c^*$ & Action $A$ & Target level $\mathcal{L}_{target}$ \\
\midrule
$[0,\,0.4)$   & --              & \textsc{Proceed}       & $\varnothing$ \\
$[0.4,\,0.5)$ & --              & \textsc{Local-Refine}  & $l_I$ \\
$[0.5,\,0.8)$ & \textsc{Method} & \textsc{Method-Refine} & $l_H \!\to\! l_I$ \\
$[0.5,\,0.8)$ & \textsc{Design} & \textsc{Design-Refine} & $l_D$ \\
$[0.8,\,1.0]$ & \textsc{Method} & \textsc{Pivot}         & $l_H$ \\
$[0.8,\,1.0]$ & \textsc{Design} & \textsc{Design-Pivot}  & $l_D$ \\
\bottomrule
\end{tabular}
\end{table}

\begin{algorithm}[ht]
\caption{Hierarchy-aware deterministic routing $\pi(c^*, s^*)$}
\label{alg:routing}
\textbf{Input:} Verified hypothesis $h^* = \langle d^*, c^*, e^*, p^* \rangle$ with criticality $s^*$\\
\textbf{Output:} Routing action $A$, target abstraction level $\mathcal{L}_{target}$
\begin{algorithmic}[1]
\IF{$s^* < 0.4$}
    \STATE \textbf{return} (\textsc{Proceed}, $\varnothing$) \COMMENT{Accept results, continue pipeline}
\ELSIF{$s^* < 0.5$}
    \STATE \textbf{return} (\textsc{Local-Refine}, $l_I$) \COMMENT{Standard reflection loop}
\ELSIF{$c^* = \textsc{Method}$}
    \IF{$s^* \ge 0.8$}
        \STATE \textbf{return} (\textsc{Pivot}, $l_H$) \COMMENT{Abandon hypothesis, regenerate}
    \ELSE
        \STATE \textbf{return} (\textsc{Method-Refine}, $l_H \!\to\! l_I$) \COMMENT{Keep hypothesis, revise method}
    \ENDIF
\ELSIF{$c^* = \textsc{Design}$}
    \IF{$s^* \ge 0.8$}
        \STATE \textbf{return} (\textsc{Design-Pivot}, $l_D$) \COMMENT{Overhaul experimental protocol}
    \ELSE
        \STATE \textbf{return} (\textsc{Design-Refine}, $l_D$) \COMMENT{Update baselines/metrics}
    \ENDIF
\ENDIF
\end{algorithmic}
\end{algorithm}

\subsection{Threshold Sensitivity}
\label{app:sensitivity}

The router exposes three hand-set constants: the severity cut-offs $0.4/0.5/0.8$ (\cref{tab:routing}), the data-sufficiency clamp floor $\kappa_0=0.6$, and the pivot budget $\texttt{MaxPivots}=2$. To check that the recovery behavior is not an artifact of these specific values, we replay the deterministic router $\pi$ on the logged diagnoses $(c^*, s^*)$ of all 11 recovered topics under perturbed thresholds. This is a decision-level analysis: it asks whether the router would have routed differently, and does not re-execute experiments.

The routing decisions are stable (\cref{tab:sensitivity}). Perturbing each severity cut-off by $\pm 0.05$ changes $0$ of the $26$ structural decisions, because the observed criticalities concentrate at $0.72$ (design) and in $[0.95, 1.0]$ (method/code), away from every boundary. The clamp floor has margin: all $11$ valid-verdict downgrades from \textsc{Pivot} to \textsc{Method-Refine} persist for $\kappa_0 \in [0.5, 0.7]$ and only begin reverting at $\kappa_0 \ge 0.8$. The pivot budget rarely binds: only ML02 issues two pivots, so $\texttt{MaxPivots} \in \{2, 3\}$ are equivalent, and only $\texttt{MaxPivots}=1$ would truncate a run. We note that the boundary insensitivity partly reflects a coarse, few-band criticality scale (the critic emits a small set of discrete severities) rather than fine-grained calibration; a finer critic is left to future work.

\begin{table}[!t]
\centering
\footnotesize
\setlength{\tabcolsep}{6pt}
\renewcommand{\arraystretch}{1.15}
\caption{Threshold-sensitivity of the router, replayed on the logged diagnoses of the 11 recovered topics ($30$ decisions; $26$ structural). Routing decisions are insensitive to the severity boundaries and to the clamp floor within $[0.5,0.7]$; only the pivot budget binds, and only for ML02.}
\label{tab:sensitivity}
\begin{tabular}{@{}lll@{}}
\toprule
\textbf{Param. (default)} & \textbf{Perturbation} & \textbf{Effect on routing} \\
\midrule
Severity cut-offs ($0.4/0.5/0.8$) & each $\pm 0.05$ & $0/26$ decisions change \\
Clamp floor $\kappa_0$ ($0.6$) & $0.5 \to 0.95$ & stable on $[0.5,0.7]$; reverts at $\kappa_0 \ge 0.8$ \\
\texttt{MaxPivots} ($2$) & $\{1,2,3\}$ & binds only ML02; $2\!\equiv\!3$ \\
\bottomrule
\end{tabular}
\end{table}

\section{Implementation Details}
\label{app:impl}
In all reported runs, the divergent attribution stage samples $K\in[3,5]$ candidate causes and retries generation when the candidates collapse to a single category or fail to cite concrete evidence from the failure context. The critic is queried independently from the generator and returns a rectified category, evidential rationale, and scalar criticality score for each candidate. The deterministic router uses the bands in \cref{tab:routing}; the maximum number of structural pivots is two, after which the system proceeds with the best available measured result or abstains if no metric-bearing run exists. The zero-metrics fast path is triggered when the execution stage emits no measured metric, no result artifact, or only infrastructure-level counts and timings.

\section{Numeric Registry and Sanitizer}
\label{app:antifab}
The numeric registry $\mathcal{R}$ is constructed from the promoted-best experiment summary and its machine-readable result artifacts. It stores measured scalar values, condition names, metric names, rounded variants, percentage variants, and arithmetic combinations explicitly allowed by the grounding manifest. During post-processing, every numeric table cell in the drafted paper is matched against $\mathcal{R}$ within a $1\%$ relative tolerance; unmatched values are replaced with the redaction sentinel \texttt{-{}-{}-}. The verifier also flags empirical drafts with no measured metrics and method or condition names absent from the registry. This mechanism checks numeric reporting only. It does not verify that prose-level claims about datasets, libraries, models, or statistical tests are supported by the executed code, which is the method-provenance gap analyzed in \cref{sec:exp-arcbench}.

\section{The AR-Eval Rubric}
\label{app:areval}
AR-Eval is the calibrated autonomous-research bar used in~\cref{sec:exp-quality}. It scores the paper produced by a fully autonomous pipeline on the axes that matter for such a system, rather than on whether the paper would be accepted at a venue. It is designed to be reported alongside a strict venue-style review, never in its place. The full instrument is given below.

\begin{rubricbox}
\textbf{Sign-inversion principle.} A venue reviewer treats every blank cell, degraded-mode banner, and missing number as a defect. For a pipeline whose explicit goal is to fail honestly rather than fabricate, that frame mis-signs the most important behavior. AR-Eval therefore credits \emph{honest abstention}. Blanking an unverified number to \texttt{-{}-{}-}, printing a degraded-mode banner, or reporting a null result is categorically better than a confident fabrication. The rubric reserves its penalties for genuine integrity failures: claims unsupported by the paper's own evidence, fabricated or ungrounded numbers, and non-executed experiments presented as real. This does not lower the bar on correctness. Internal self-contradiction, fabrication, and non-execution are penalized at least as hard as a venue reviewer would penalize them.

\tcblower
\textbf{Dimensions.} Each dimension is scored on an integer scale from 1 (poor) to 5 (excellent), with 3 adequate.

\textbf{AC, Autonomy and Completion.} Did the pipeline run end to end to a complete, compiling paper without human intervention? A 5 is a complete paper with all standard sections that compiles and references its figures and tables. A 3 is complete but with structural rough edges, such as placeholder captions or minor missing sections. A 1 is no paper, a blocked run, or a document truncated mid-text.

\textbf{EG, Empirical Grounding.} Are the reported results backed by real executed experiments with real measured metrics? A 5 traces every result to an executed run with clear provenance. A 3 has partial real metrics, with some conditions missing but the present ones real. A 1 is simulated or formulaic data, or no metrics at all. Honestly blanked cells do not lower EG on their own, since EG measures whether real measurement happened, not whether every cell is filled.

\textbf{RI, Reporting Integrity.} Does the artifact represent its evidence honestly? This is the sign-inverted dimension. It is \emph{raised} by blanking unverified numbers, by degraded-mode or requires-verification banners, by honestly reported null or negative results, and by calibrated claims that match the evidence. It is \emph{lowered} by prose, abstract, or conclusion numbers with no matching populated table cell, by fabricated or ungrounded numbers, by claims that exceed the presented evidence, and by bolding a blank cell as the best value. A 5 is fully self-consistent, with every stated number backed and every gap honestly marked. A 3 is mostly consistent, with at most one minor prose-to-table mismatch. A 1 has pervasive prose-to-table contradictions or fabricated values.

\textbf{MS, Methodological Soundness (relative).} Is the experimental design reasonable for an autonomous run? We judge relative to what is achievable without human intervention, so we do not require five seeds and confidence intervals to pass. Blatant design holes still cost, such as a single-seed run with strong causal claims, untuned baselines presented as decisive, or no control. A 5 is a clean controlled design with multiple seeds and appropriate baselines and metrics. A 3 is a reasonable design with acknowledged limits and no overclaim. A 1 is a design that cannot support any stated conclusion. This dimension is our least reliable across judges and should be read as low-confidence (\cref{sec:exp-quality}).

\textbf{SC, Scientific Coherence and Contribution.} Is there a coherent question and a real finding, with grounded related work? A negative or small finding is acceptable when it is honestly framed and useful. A 5 has a clear question, a genuine finding, and is well situated. A 3 is coherent but thin or incremental. A 1 has no discernible question or finding.

\smallskip
\textbf{Aggregation and bands.} Let $\overline{D} = \tfrac{1}{5}\sum_{i=1}^{5} D_i$ be the mean dimension score. The overall score is tied to the five dimensions rather than set independently:
\[
    \text{AR-Eval} = \operatorname{round}\!\big(2\,\overline{D}\big) + \Delta, \qquad \Delta \in \{-1, 0, +1\},
\]
where $\Delta$ is a single holistic adjustment of at most one point in either direction, applied for an overriding strength or defect. The verdict bands are: 8 to 10, a strong autonomous artifact that is complete, grounded, self-consistent, and reports a real finding; 6 to 7, a credible autonomous artifact that is complete and honest but whose finding is thin or whose limits are significant; 4 to 5, partial, a run with real gaps in grounding or integrity; and 1 to 3, incomplete or unreliable, meaning blocked, fabricated, or self-contradictory. To keep each score auditable, the judge also returns whether honest abstention is present and the list of genuine, bar-independent integrity defects that cost points, and emits a final line of the form \texttt{AR-SCORE | AC=x | EG=x | RI=x | MS=x | SC=x | overall=x/10 | band=<band>}.

\smallskip
\textbf{Protocol and reporting caveats.} Each paper is scored blind by an independent reviewer that sees only that paper's text. We run two judge models, Claude Opus 4.8 and Claude Sonnet 4.6, and report their agreement in \cref{sec:exp-quality}. We attach four caveats: \textbf{(i) dual-standard}, report AR-Eval next to a strict venue-style score, since the contrast is the finding, not the AR-Eval number alone; \textbf{(ii) judge disclosure and self-preference}, disclose the judge model and the full rubric, mitigate self-preference and verbosity bias with at least two judge models or multiple samples, and validate against human ratings on a held-out subset, reporting Spearman $\rho$ or Cohen's $\kappa$; \textbf{(iii) blind and paper-only}, the reviewer sees only the paper text, never the execution logs or the arm label; and \textbf{(iv) not an acceptance claim}, a high AR-Eval score means a credible autonomous research artifact, not a paper publishable at a given venue. We additionally validate system ordering through the blind three-reviewer study on the six ML topics described in Section 4.4. Validation outside ML remains outstanding.

\end{rubricbox}

\section{Case Study: Real Metrics Do Not Guarantee a Better Paper (B07)}
\label{app:b07}
This case study follows the writer-to-sanitizer coordination failure of~\cref{sec:exp-integrity} through one deliverable end to end. B07 is the clearest case of the gap between producing results and reporting them. Under \textsc{SAGE} the experiment executed and measured real flux-balance results: an \emph{E.\ coli} growth rate of $0.4678/\mathrm{h}$, per-method active-reaction counts, and runtimes across FBA, pFBA, and FVA. Its experiment is therefore metrics-bearing. The paper nonetheless scores $5/10$, below the \textsc{Baseline}'s $6/10$ (\cref{fig:quality}). The cause is internal, not scientific. The writer states these measured numbers in the abstract, body, and figure captions, while the sanitizer blanks the matching result-table cells to \texttt{-{}-{}-}, and one table bolds a blanked cell as the column best (\cref{fig:b07cases}). A reviewer reading confident prose against an abstaining table penalizes the contradiction, so reporting integrity falls to $2/5$. The \textsc{Baseline} paper, an honestly-framed protocol that measured nothing and claims nothing, draws no such penalty. A paired re-evaluation of both arms in a single round confirms the ordering, so it is not a cross-round artifact.

\begin{figure}[t]
\centering
\begin{tcbraster}[raster columns=2, raster column skip=8pt, raster equal height=rows]
  \begin{tcolorbox}[colback=orange!12, colframe=orange!70!black, coltitle=white, fonttitle=\bfseries\small, title={Problem 1: prose contradicts the table}, boxrule=0.6pt, arc=3pt, fontupper=\small]
  The abstract and body assert measured numbers: an \emph{E.\ coli} growth rate of $0.4678/\mathrm{h}$, an active-reaction count of $49$, and an FVA runtime of $4.74$\,s (about $4{,}700\times$ FBA). The matching cells in Tables~2 and~4 are blanked to \texttt{-{}-{}-}. The reader meets confident numbers the artifact itself declines to report.
  \end{tcolorbox}
  \begin{tcolorbox}[colback=red!10, colframe=red!70!black, coltitle=white, fonttitle=\bfseries\small, title={Problem 2: a blank cell bolded as ``best''}, boxrule=0.6pt, arc=3pt, fontupper=\small]
  Table~2 marks the \texttt{n\_active} column winner as \textbf{\texttt{-{}-{}-}}, and Table~4 prints the runtime ratio as \texttt{4,-{}-{}-}$\times$. A redaction sentinel is typeset as the best value and as a half-formed number, which is neither honest abstention nor a measured result.
  \end{tcolorbox}
\end{tcbraster}
\caption{Two reporting-integrity defects in the B07 \textsc{SAGE} deliverable, both from the writer--sanitizer coordination failure of~\cref{sec:exp-integrity}: the writer emits numbers the sanitizer then blanks. These defects, not the science, cap the paper's reporting integrity at $2/5$.}
\label{fig:b07cases}
\end{figure}

We read B07 as a correct evaluation of a fixable defect. The rubric rewards an honest empty paper over a self-contradictory one, which is the intended behavior. The measured values exist in the run and would survive verification if the writer populated table cells from the verified registry rather than restating them only in prose. Closing that loop is the single change that would let B07's real science reach the page.

\section{Limitations and Future Work}
\label{app:limitations_future}

While \textsc{SAGE} demonstrates that structured failure attribution significantly improves the reliability of autonomous agents, our study presents certain limitations that also serve as a roadmap for future exploration in full auto-research:

\paragraph{1. Computational Budgets versus Unbounded Discovery.} 
\textsc{SAGE} is fundamentally built on a continual, self-correcting paradigm. In an ideal scientific scenario, the agent would autonomously judge exactly when to terminate the MHFA loop based on pure scientific convergence or definitive falsification. However, to manage the practical economic costs of LLM API calls and sandbox compute during our experiments, we imposed strict, hard-coded iteration limits (e.g., the \texttt{MaxPivots} budget). As inference costs continue to decrease, we envision future ``full auto-research'' frameworks operating without these artificial bounds. Unconstrained by predefined budgets, agents will be able to execute as many diagnostic cycles as necessary to thoroughly explore dead ends, overcome deep structural flaws, and ultimately uncover genuinely novel scientific findings.

\paragraph{2. Evaluation Scale and Cross-Domain Generalization.} 
Although our 12-topic subset was carefully curated to span five diverse scientific domains, it remains a relatively small-sample validation. Transitioning this framework to hundreds of open-ended research topics across wet-lab, physical, and broader computational sciences is a necessary next step. We hope that expanding the evaluation scale will not only stress-test our hierarchy-aware router but also inspire the broader community to build upon the \textsc{SAGE} architecture. Ultimately, our goal is to shift the field away from brittle and hallucination-prone pipelines by constructing a trustworthy and rigorously grounded paradigm for autonomous research. This reliable foundation will be essential for driving future breakthroughs in critical areas, particularly in medical research and the discovery of new chemical materials.

\paragraph{3. Expanding the Hypothesis Space with External Knowledge.} 
Currently, the Divergent Causal Generation stage of MHFA relies on the internalized, parametric knowledge of the foundation models to diagnose failures. For highly specialized or completely novel scientific domains, the model might occasionally lack the niche expertise required to identify obscure structural flaws. A highly promising future direction is to integrate Retrieval-Augmented Generation (RAG) into the MHFA pipeline. By empowering the agent to autonomously query external scientific literature, experimental databases, or codebase documentation during the diagnostic phase, the system could formulate even more sophisticated and domain-specific hypotheses, pushing the boundaries of autonomous scientific discovery to unprecedented levels.

\paragraph{4. Foundation Model Scaling and Fair Comparison.} 
To ensure a fair comparison, we used the same foundation model for all systems in our evaluation. This control ensures that \textsc{SAGE}'s performance gains come entirely from our MHFA architecture, rather than from using a stronger LLM. However, the system's absolute performance remains tied to the capabilities of the underlying model. Because \textsc{SAGE} is a model-agnostic framework, it can easily integrate advanced models like Claude Ultracode. By leveraging such a model's native sub-agent orchestration capabilities, \textsc{SAGE} could divide the failure diagnosis process into finer-grained, multi-angle tasks. We expect that this synergy will yield even more reliable and trustworthy scientific artifacts as foundation models continue to evolve.


\end{document}